\documentclass[10pt,twocolumn,letterpaper]{article}
\pdfoutput=1 

\usepackage{cvpr}
\usepackage{times}
\usepackage{epsfig}
\usepackage{graphicx}

\usepackage[sort&compress,numbers]{natbib}

\usepackage{amsmath}
\usepackage{amssymb}

\usepackage{booktabs}
\usepackage{multirow}
\usepackage{adjustbox}
\usepackage{stackengine} 

\usepackage{balance,flushend}
\usepackage{subcaption}

\usepackage{xcolor}
\newcommand{\cameraready}[1]{#1}

\usepackage[breaklinks=true,hidelinks,bookmarks=false]{hyperref}

\cvprfinalcopy 

\begin{document}

\title{Adaptive Multiplane Image Generation from a Single Internet Picture}

\author{%
  \begin{tabular}{@{}c@{}}
    Diogo C. Luvizon$^1$\thanks{\scriptsize This work was funded by Samsung Eletr\^{o}nica da Amaz\^{o}nia Ltda., through the project ``Parallax Effect'', within the scope of the Informatics Law No. 8248/91.} \,
    Gustavo Sutter P. Carvalho$^{1,2}$ \,
    Andreza A. dos Santos$^3$ \,
    Jhonatas S. Concei\c{c}\~{a}o$^3$ \\
    Jose L. Flores-Campana$^3$ \,
    Luis G. L. Decker$^3$ \,
    Marcos R. Souza$^3$ \,
    Helio Pedrini$^3$ \\
    Antonio Joia$^1$ \,
    Ot\'{a}vio A. B. Penatti$^1$
  \end{tabular}\\
  $^1$AI R\&D Lab, Samsung R\&D Institute Brazil, Campinas, SP, 13097-160, Brazil\\
  $^2$ICMC, University of São Paulo (USP), São Carlos, SP, 13566-590, Brazil \\
  $^3$Institute of Computing, University of Campinas (UNICAMP), Campinas, SP, Brazil, 13083-852 \\
}

\maketitle

\begin{abstract}
In the last few years, several works have tackled the problem of novel view synthesis from stereo images or even from a single picture. However, previous methods are computationally expensive, specially for high-resolution images. In this paper, we address the problem of generating a multiplane image (MPI) from a single high-resolution picture. We present the adaptive-MPI representation, which allows rendering novel views with low computational requirements. To this end, we propose an adaptive slicing algorithm that produces an MPI with a variable number of image planes. We present a new lightweight CNN for depth estimation, which is learned by knowledge distillation from a larger network. Occluded regions in the adaptive-MPI are inpainted also by a lightweight CNN. We show that our method is capable of producing high-quality predictions with one order of magnitude less parameters compared to previous approaches. \cameraready{The robustness of our method is evidenced on challenging pictures from the Internet.}
\end{abstract}

\section{Introduction}

Novel view synthesis from a single image is a very challenging problem that has gained attention from the computer vision community in the last few years. Despite the challenges involved in this task, for instance, estimating depth and generating color information on occluded regions, novel view synthesis unlocks a broad range of applications related to 3D visual effects from a single image, augmented reality systems, among many others.

Traditionally, methods for new view synthesis rely on stereo vision for obtaining both depth information and color inpainting~\cite{Penner_2017_TOG, zhou2018stereo}. Among these methods, multiplane images (MPI)~\cite{szeliski1999stereo} has frequently been used to encode the scene. Contrarily to other representations, such as point clouds or 3D meshes, multiplane images are easily stored with traditional image compression algorithms and can be rendered efficiently on embedded systems, such as in smartphones or Smart TVs, even for high-resolution images.

\begin{figure}[!t]
\centering
\includegraphics[width=0.95\columnwidth]{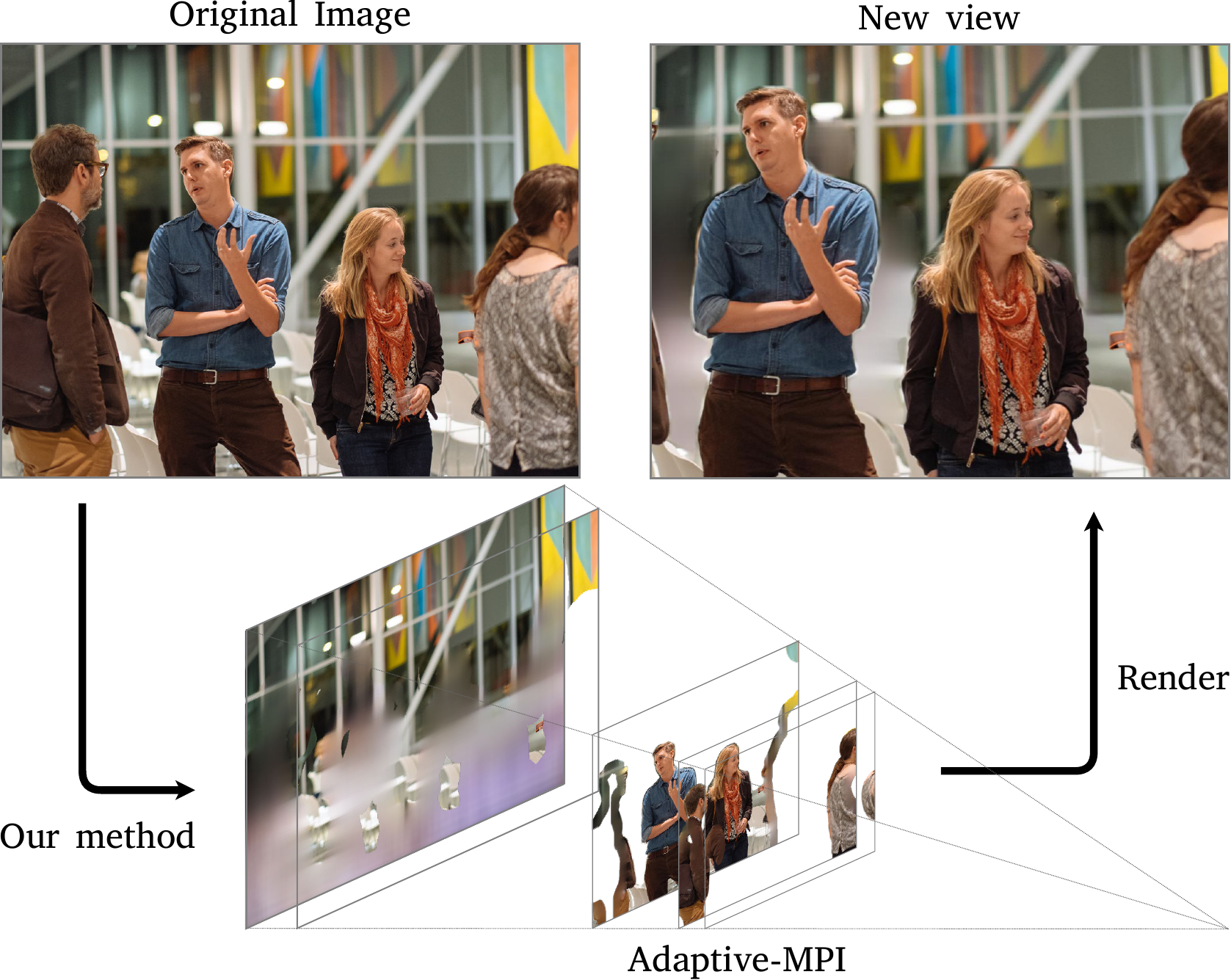}
\caption{Our method predicts an adaptive multiplane image representation from a single picture, which can be easily rendered to produce new views of the scene.
}
\label{fig:intro}
\end{figure}

Learning an MPI representation from a single image is a much more challenging problem, since depth and occluded colors have to be inferred from a flat image. Very recently, Tucker and Snavely~\cite{Tucker_2020_CVPR} proposed a framework for learning an MPI from a single-view image. However, their method is still dependent on multiview (stereo) data during training. This fact hinders the robustness and applicability of the method, since such data is often collected from a moving camera (e.g., images from RealEstate10K dataset), therefore, being limited to static scenes. Contrarily, in our method we learn an MPI representation from single-view and unconstrained pictures from the Internet. As a result, our method benefits from a strong generalization capability. Another drawback of previous methods based on MPI is their high computational cost, since a high number of image planes are often used to achieve satisfactory results~\cite{srinivasan2019pushing}.

Considering the limitations of previous methods, we propose a new approach to generate an \textit{adaptive multiplane image} (adaptive-MPI) from a single image. As shown in Figure~\ref{fig:intro}, differently from  traditional MPI, the adaptive version automatically selects the number of layers and the distance between them in order to better represent the content of the scene. Consequently, our method is able to produce results comparable to the state of the art while requiring very few image planes and running in a few seconds even for high-resolution images. The main contributions of our work are listed as follows. \textit{First}, we propose a new method to learn an adaptive-MPI from single images from the Internet. \textit{Second}, we propose a lightweight CNN for depth estimation and a robust distillation method for transfer learning from a larger network in high-resolution. \textit{Third}, we propose an inpainting strategy that handles MPI with a variable number of layers. Our approach is evaluated for depth estimation with respect to our \textit{teacher} network and on view synthesis on challenging pictures extracted from the Internet.


\section{Related work}
\label{sec:relatedwork}

In this section, we review some of the most relevant works related to our method.

\subsection{Single-image depth estimation}

Estimating depth from a single image is an ill-posed problem due to scale ambiguity. With the advent of deep learning, predicting depth maps has become possible for well conditioned
environment~\cite{eigen2014depth}, such as indoor scenes. However, for reducing the epistemic uncertainty related to the
task~\cite{NIPS2017_7141}, an enough amount of data is required for more general cases. To handle this issue, methods based on stereo
matching have emerged in the last few years~\cite{Godard_2017_CVPR, Tosi2019CVPR}. Such approaches rely only on well calibrated pairs of images for training, which is easier to capture than precise depth maps. However, learning from stereo pairs is not as effective as learning directly from ground-truth depth~\cite{FuCVPR2018}.

More recently, impressive results have been achieved by exploring stereo data from the Web~\cite{XianCVPR2018}, structure from motion (SfM) using Internet pictures~\cite{li2018megadepth}, and from YouTube videos (mannequin challenge)~\cite{li2019learning}. The main advantage of these methods is that they benefit from a high variety of data. However, when multiple datasets with different scales and data distributions are mixed together, learning a common representation becomes a challenge. Ranftl et al.~\cite{Ranftl2020} proposed to solve this problem with a scale- and shift-invariant loss, that allows the network learning from multiple datasets simultaneously.

Considering the discussed previous work, we can notice that a key factor for robust depth estimation is training data. Nevertheless, some datasets may not be readily available or may require customization that makes their use difficult. Therefore, our main contribution in depth estimation is a robust distillation process for learning a lightweight model
capable of generating high-quality and high-resolution depth maps, as demonstrated through our experiments.

\subsection{View synthesis from a single view}

Realistic view synthesis has gained great relevance in the areas of computer and robotic vision. Recent works have used deep learning-based approaches, which generally predict representations such as 3D meshes~\cite{liu2019soft} or point clouds~\cite{wiles2020synsin}, while for more complex images, representations such as MPI~\cite{szeliski1999stereo} and LDI~\cite{shade1998layered} have been used. Until recently, methods in the literature have only dealt with calibrated stereo images~\cite{zhou2018stereo,flynn2019deepview, srinivasan2019pushing} and used fully-convolutional deep networks to predict an MPI representation, which is then rendered to extrapolate novel views.

Currently, there are still few studies that predict new views from a single image, such as in~\cite{Tucker_2020_CVPR}. In this case, the authors generated the training data for depth supervision using Simultaneous Localization and Mapping (SLAM) and SfM algorithms from videos with static content, while the network is optimized for both depth estimation and new view synthesis. \cameraready{Image segmentation has also been explored to simulate motion parallax effect~\cite{pinto2020parallax}}. Other methods propose to learn an LDI relying on intermediate tasks, such as depth and segmentation maps~\citep{dhamo2019peeking,dhamo2019objectdriven}, or even from pairs of stereo images~\cite{tulsiani2018layer}. Very recently, Shih et al.~\cite{shih20203d} proposed a framework for new view synthesis based on monocular depth estimation and on a cascade of depth and color inpainting. \cameraready{Despite the high-quality results, this method has an iterative process that becomes prohibitive for high-resolution images on computational restricted devices.}

Similar to the work proposed by~\cite{Tucker_2020_CVPR}, we predict an MPI from a single image. However, the planes of our MPI are defined in an adaptive way, depending on the input image, which is more suitable for view synthesis from a small number of image planes. In addition, our method does not rely on stereo data for training and performs efficiently due to our lightweight models.

\subsection{Image inpainting}

Several inpainting techniques have been proposed to fill missing or occluded parts of an image. These methods can be divided into traditional diffusion-based~\cite{2000bertalmio_DiffusionInpainting, 2001Bertalmio_diffusionBased} and patch-based~\cite{2003drori_patchedInpainting} approaches, and more recent GAN-based methods~\cite{pathak2016context, SIGGRAPH2017_LocalGlobalConsistent, yu2019free}. Differently from classic image inpainting, multiplane images require a more sophisticated process, since only the background or specific borders in the image layers need inpainting. Since we are not considering the case of stereo images for training, we propose an approach to learn an inpainting CNN for MPI layers based on single images and their respective depth maps.

\section{Proposed method}
\label{sec:method}


\subsection{Dataset preparation}

The goal of our method is to produce an adaptive-MPI representation from unconstrained images. Publicly datasets for novel view synthesis are frequently based on static scenes, i.e., collected from videos without people or animals, such as RealEstate10K~\cite{zhou2018stereo},
limited to macro pictures captured by plenoptic
cameras~\cite{LearningViewSynthesis, srinivasan2017learning}, or restricted to a specific domain~\cite{geiger2013vision}. Available depth estimation datasets are also limited to specific domains (MegaDepth~\cite{li2018megadepth}) and low resolution images (NYU-Depth V2~\cite{silberman2012indoor}).

\begin{figure}[!htbp]
\centering
    \includegraphics[height=1.50cm]{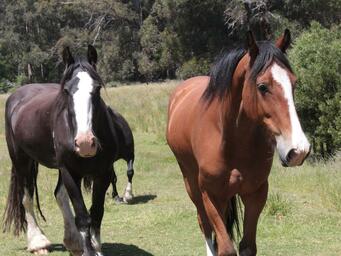}
    \includegraphics[height=1.50cm]{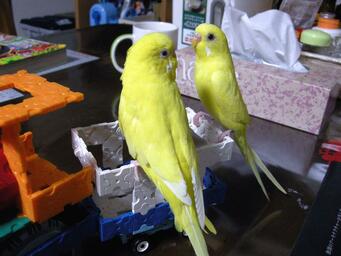}
    \includegraphics[height=1.50cm]{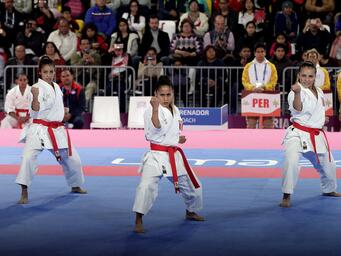}
    \includegraphics[height=1.50cm]{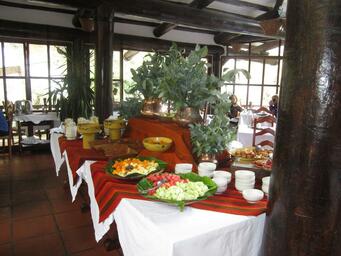}
    \includegraphics[height=1.50cm]{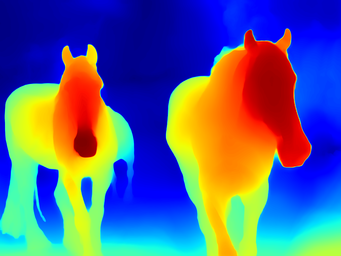}
    \includegraphics[height=1.50cm]{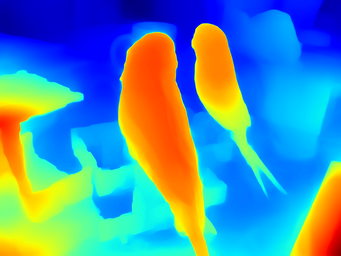}
    \includegraphics[height=1.50cm]{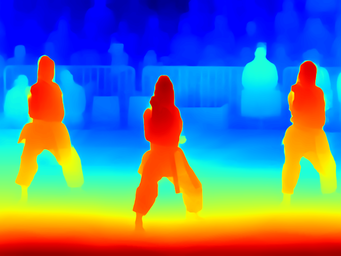}
    \includegraphics[height=1.50cm]{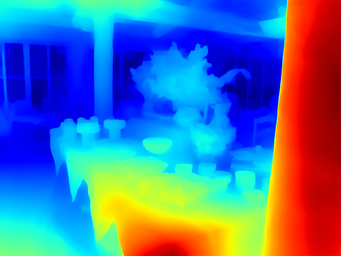}
\caption{Samples from our dataset and the pseudo ground truth generated by our ensemble approach.
}
\label{fig:dataset}
\end{figure}

Nevertheless, the Internet is a massive source of high-resolution image data, with scenes of different modalities, including landscapes, portraits, groups of people, animals, food and vehicles, frequently on variable conditions of lightning. This abundance of image data could be used to train a robust inpainting algorithm, as well as to distill the knowledge from depth estimation methods. With this in mind, we collected a dataset of high-resolution Internet pictures, which is used for teaching our lightweight CNN models, both for depth estimation and inpainting tasks. Some examples of images from our dataset are shown in Figure~\ref{fig:dataset}.

\begin{figure*}[!htbp]
\centering
\includegraphics[width=0.98\textwidth]{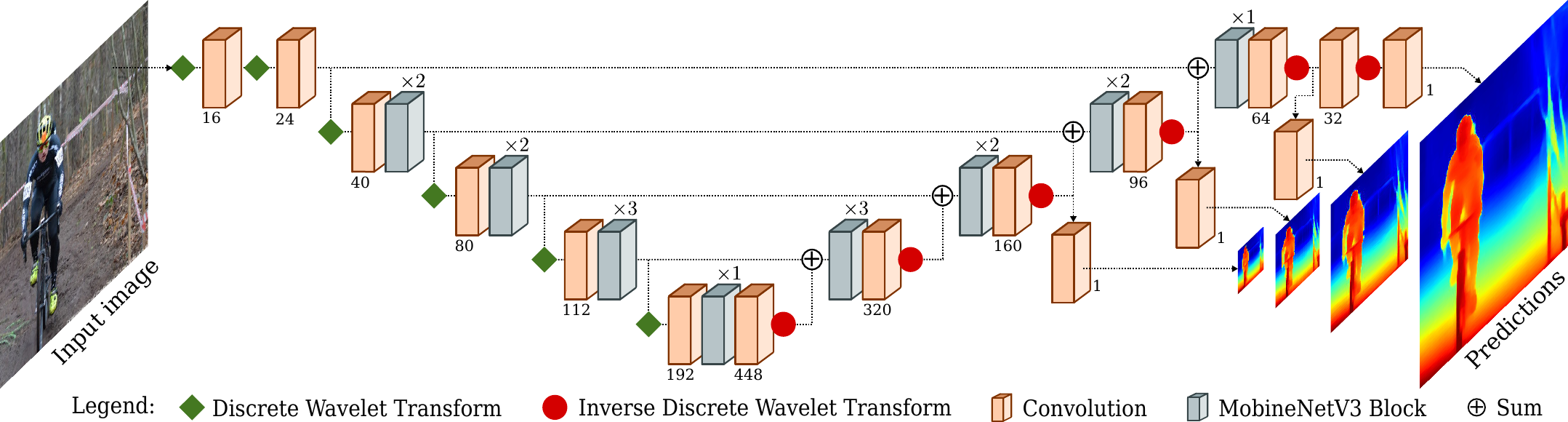}
\caption{Depth estimation network (\textit{lite}) based on MobileNet V3 blocks. Input and feature maps are downscaled by Discrete Wavelet Transforms (DWT) and upscaled by Inverse DWT. Supervision at intermediate resolutions are applied to enforce multi-resolution learning.
}
\label{fig:depthnetwork}
\end{figure*}

\subsubsection{Data collection}

We collected 100K images from Flickr by performing queries based on 16 different keywords:
\textit{america}, \textit{animals}, \textit{asia}, \textit{brazil}, \textit{city}, \textit{civilization}, \textit{europe}, \textit{food}, \textit{indoor}, \textit{landscape}, \textit{night}, \textit{objects},
\textit{people}, \textit{show}, \textit{sports}, and \textit{vehicle}. The queries were made for high-resolution images with no restrictions in the rights of use. All the collected files were filtered to remove duplicated pictures and to correct rotations. We split the remaining files as 95K training and 3K evaluation images.

\subsubsection{Distilling depth knowledge}
\label{sec:disitllingdepth}

Recent depth estimation methods~\cite{Ranftl2020, li2018megadepth} are capable of predicting high quality depth maps. However, these models are computationally expensive, specially for high-resolution images. Since our approach focuses on efficiency, we propose to distill the knowledge from MiDaS~\cite{Ranftl2020} by using our collected pictures from the Internet.

In knowledge distillation, an effective approach is to train the \textit{student network} based on an ensemble of predictions from the \textit{master network}~\cite{hinton2015distilling}. In our method, we performed a 10$\times$ ensemble by considering predictions in five different resolutions (squares with rows and columns with $512$, $768$, $1024$, $1600$, and $1920$ pixels) and horizontal flipping. All the intermediate predictions were then normalized and resized with bilinear interpolation to the highest resolution, which are then combined by the median value at each pixel position, resulting in our \textit{pseudo ground truth} depth maps. Figure~\ref{fig:dataset} shows some examples of depth maps generated by this approach. \cameraready{For simplicity and efficiency during training, out method does not rely on intermediate feature supervision.} Similarly to previous works, we use the term \textit{depth map} interchangeably for \textit{disparity map}.

\subsection{Efficient depth estimation}

Considering our objective to produce high-resolution effects efficiently, we propose a lightweight CNN for depth estimation. Previous methods for depth estimation are frequently based on U-Net~\cite{li2019learning} architectures. The drawback of traditional U-Net models are their high number of parameters and high memory usage, specially for high-resolution images. Instead, we propose to adapt EfficientNet~\cite{Mingxing2019EfficientNet} to perform depth estimation.

Specifically, we integrated the lightweight MobileNet V3 block~\cite{Howard_2019_ICCV} with Discrete Wavelet Transforms (DWT), as shown in Figure~\ref{fig:depthnetwork}. We use DWT and Inverse DWT in a similar way to Luo et al.~\cite{Luo_2020_CVPR}, i.e., for downscaling and upscaling feature maps, respectively. The DWT has the effect of decomposing an input tensor of shape $H\times{W}\times{C}$ into a set of approximation coefficients (AC) and detailed coefficients (DC), which represent the time-frequency decomposition of the input signal. By concatenating the AC and DC components, we recover a tensor with shape $H/2\times{W/2}\times{4C}$. This allows our model to benefit from skip connections with lower resolutions compared to traditional U-Nets and to preserve high frequency information throughout the full network. In addition, we perform multi-scale depth supervisions to enforce learning from multi-resolution signals. Our design decisions are evaluated in the ablation studies.

\subsection{Adaptive multiplane image slicing}

The goal of the adaptive multiplane image slicing algorithm is to compute a small set of image planes that represents a 3D scene, while reducing the possible artifacts on novel views. To this end, we propose \cameraready{a slicing algorithm based on depth information. This algorithm selects} thresholds between the image planes from regions with high discontinuity in the depth map, \cameraready{assuming that the scene has some depth salience, such as an object in the foreground.}

\begin{figure}[!htbp]
\centering
\stackunder[5pt]{\includegraphics[width=0.155\textwidth]{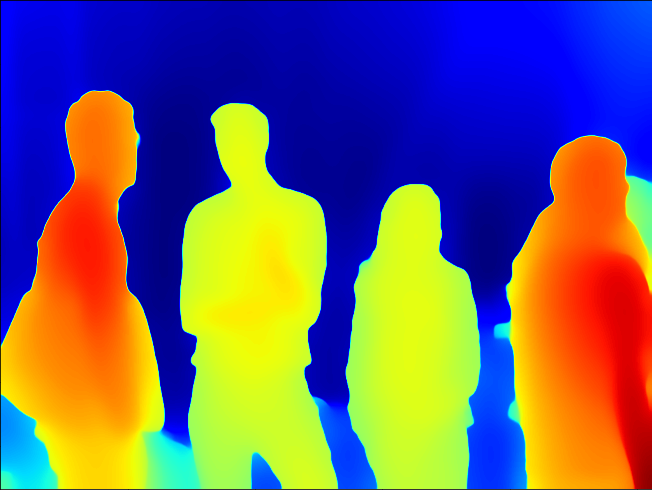}}{Filtered depth}
\stackunder[5pt]{\includegraphics[width=0.155\textwidth]{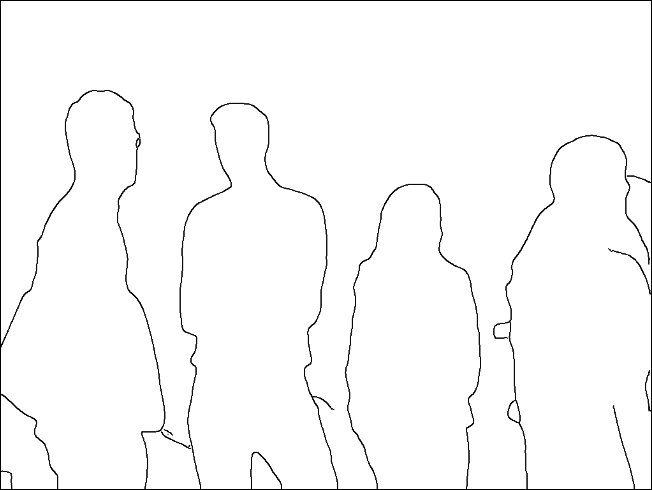}}{Canny edges}
\stackunder[5pt]{\includegraphics[width=0.155\textwidth]{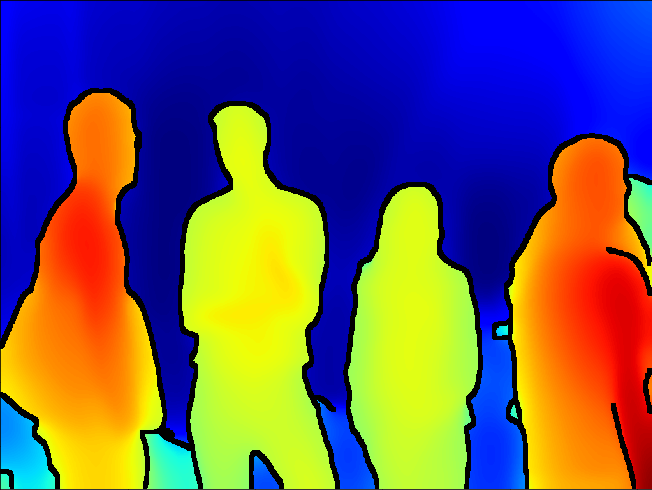}}{Eroded borders}
\caption{Pre-processing of depth maps: proposed steps allow better delineation of object edges.}
\label{fig:preprocess}
\end{figure}

\begin{figure}[!htbp]
\centering
\includegraphics[width=0.475\textwidth]{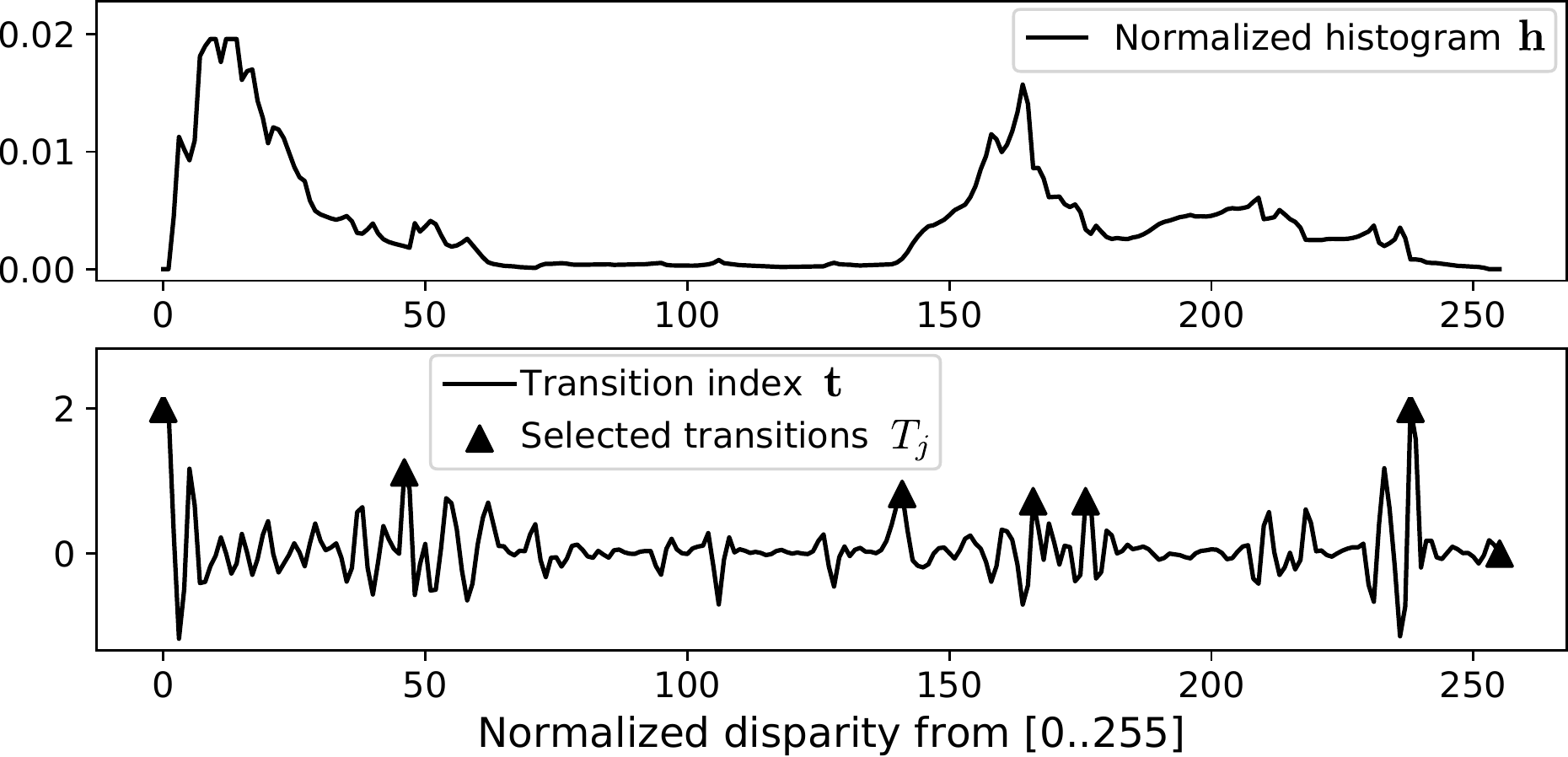}
\caption{Local and spaced maximum values in $\textbf{t}$ define the transitions $T_j$ between image planes. The first and last index of $\textbf{t}$ are always selected as boundaries. Note that no transition was selected between 46 and 141 due to the low density in this interval.}
\label{fig:slicingmethod}
\end{figure}

Given a depth map of a scene, we first pre-process it in order to enhance depth discontinuities. In this pre-processing stage, depth values are normalized to the interval $[0..255]\in\mathbb{Z}^+$. Then, we apply a bilateral filter~\cite{tomasi1998bilateral} followed by a Canny edge detector. The borders in the depth map are expanded by a morphological erosion. The effect of this process is shown in Figure~\ref{fig:preprocess}. From the resulting depth map, we compute a \textit{transition index}, defined by:
\begin{equation}
    \textbf{t}=\frac{\nabla^2{\textbf{h}}}{\max(\epsilon, \textbf{h})},
\end{equation}
\noindent where $\textbf{h}$ is the normalized histogram of the resulting depth map and $\epsilon$ is set to $0.001$ to avoid division by zero. Figure~\ref{fig:slicingmethod} shows the normalized histogram and the transition index for a given depth map.

A high value in $\textbf{t}$ corresponds to a valley or a discontinuity in the depth map, since it depends on the second derivative of $\textbf{h}$. Therefore, the local maximum values from $\textbf{t}$ are used to select the threshold (transition) between two image planes. Once a transition $t_i$ is selected, the corresponding value of $i$ is stored in a vector $T$ and its neighbors $\{i-\xi, \dots, i+\xi\}$ in $\textbf{t}$ are set to zero. This prevents from selecting transitions too close in the depth. In our implementation, we set $\xi=8$. This process of transition selection repeats until no more values in $\textbf{t}$ are above a threshold (set to $0.1$ in our method), or a maximum number of selected transitions is reached. At the end of this process, we have $N+1$ selected transitions, where $N$ is the number of image planes. Finally, each image plane $\textbf{I}_j$ is formed by the pixels lying in the depth interval $\{\textbf{h}_{T_j},\dots,\textbf{h}_{T_{j+1}}\}$, where $j=\{0, \dots, N\}$ is the index of the selected transitions. The position of $\textbf{I}_j$ on the $Z$ axis is computed as the average of the depth values lying in the same interval. Thus, the number of image planes and the depth of each layer is adapted accordingly to the depth information of the scene.

\subsection{Multiplane image inpainting}

Once the image planes are computed by the adaptive slicing algorithm, some regions may require image inpainting to fill the holes on the generated novel views. A naive approach would be to inpainting after rendering. However, this would be prohibitive to most of the real-time applications, since inpainting is computationally expensive and cannot be performed on current embedded devices in real-time for high-resolution images. Instead, we propose to inpaint the adaptive-MPI representation before rendering, which means that no additional post-processing will be required during novel views generation.

Differently from classic inpainting tasks based on regular regions~\cite{cai2020piigan} or on small free-form regions~\cite{yu2019free}, in our case, we have to inpaint image planes that frequently contain very sparse visual information. In addition, the parts of the image that require inpainting are generally thin margins that follow the contour of the available color information. For this reason, we trained a lightweight inpainting CNN\footnote{The CNN architecture for inpainting was inspired in~\cite{yu2018generative} but with a smaller number of filters and convolutions.} for generating pixels in a masked region that is computed based on morphological operations.

\begin{figure}[!htbp]
\centering
\includegraphics[width=0.46\textwidth]{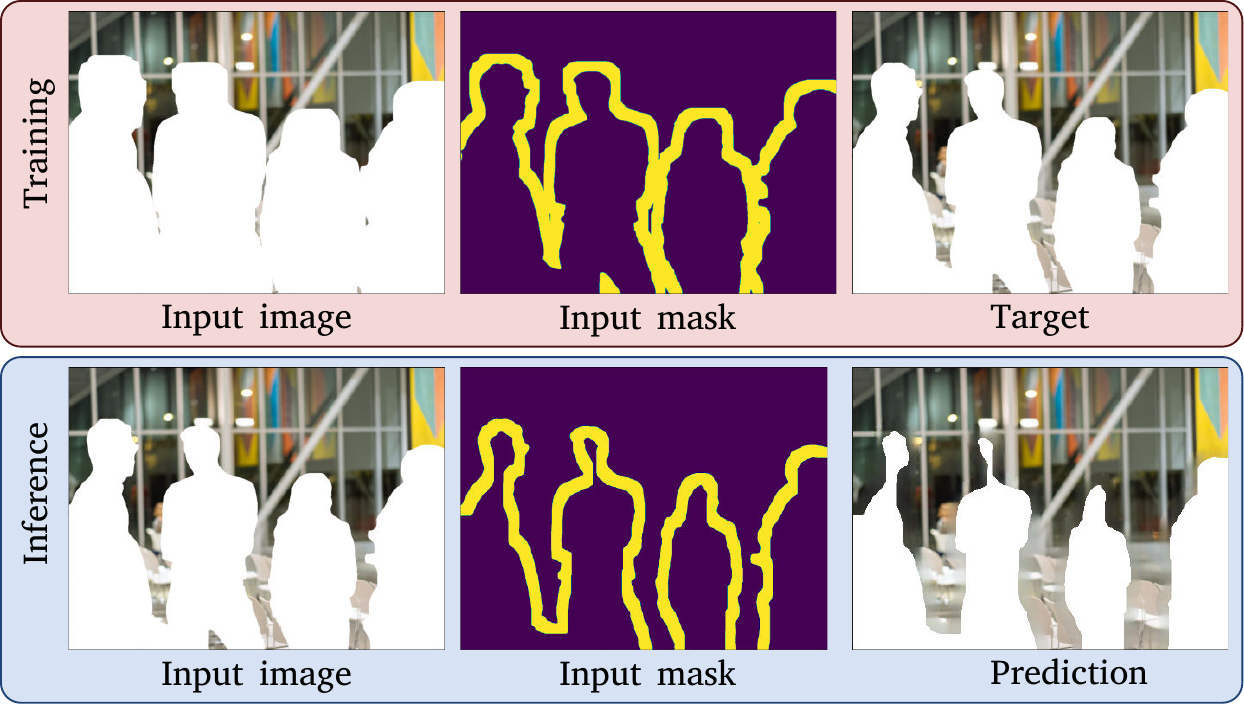}
\caption{Illustration of the proposed inpainting process during training and inference. The goal of the inpainting is to generate pixels to extend the background image (input) according to a margin defined by morphological operations (input mask).}
\label{fig:inpaintingmethod}
\end{figure}

Since our method is trained on single pictures from the Internet, therefore, no ground truth is available for occluded regions, we propose to train the inpainting model as follows: given an image layer computed by the adaptive slicing algorithm, we perform a morphological erosion in its mask in order to remove a border of pixels following the image contour. Then, the resulting image is concatenated with a mask corresponding to the removed pixels and fed to the network. The model is supervised with the original image layer (target). This process is illustrated in Figure~\ref{fig:inpaintingmethod} (top).
During inference (Figure~\ref{fig:inpaintingmethod}, bottom), the input mask is computed based on the dilated border from the current image layer, and both are fed to the network. The prediction is then used for inpainting. Transparent regions in the MPI are not considered in this process, such that pixels are generated on occluded regions only.

\section{Experiments}
\label{sec:experiments}

In this section, we present the implementation details of our method, as well as an experimental evaluation considering \textit{depth estimation} and \textit{view synthesis}, for both quantitative and qualitative results.

\subsection{Implementation details}

To train the depth estimation network, we used a composed loss function based on two terms, data and gradient losses:
\begin{equation}
    \mathcal{L}_{\textit{data}} = \frac{1}{N}\sum_{i=0}^{N}{|\hat{y} - y|}
\end{equation}
\begin{equation}
    \mathcal{L}_{\textit{grad}} = \frac{1}{N}\sum_{i=0}^{N}\sum_{s=0}^{3}{|\nabla_x(\hat{y}_s - y_s)| + |\nabla_y(\hat{y}_s - y_s)|},
\end{equation}
\noindent where $\hat{y}_s$ and $y_s$ are the predicted and the target values, considering a downscaling factor of $2^s$. Similarly to~\cite{Ranftl2020}, we normalize the target disparity maps between 0 and 10. The final depth loss is defined by
\begin{equation}
    \mathcal{L}_{\textit{depth}} = \mathcal{L}_{\textit{data}} + \alpha\mathcal{L}_{\textit{grad}},
    \label{eq:depthloss}
\end{equation}
\noindent where $\alpha$ is set to $0.5$ in our experiments. This loss is applied individually to each output of our model (see Figure~\ref{fig:depthnetwork}). At inference time, we use only the last prediction. The network was optimized with stochastic gradient descent and the Adam optimizer. Learning rate was initially set to $0.001$ and reduced by a factor of $0.2$ when validation plateaus. We trained the network with the images of our dataset rescaled to $1024\times768$ pixels, for about 1.2 million iterations using 6 images per batch.

For training the inpainting network, we randomly selected image layers computed by our adaptive slicing algorithm and included on it the respective layers farther in the depth range, resulting in one single image containing the pixels farther than a given depth threshold. The resulting image was then pre-processed as illustrated in Figure~\ref{fig:inpaintingmethod} and used for training. In the morphological dilation, we used a $3\times3$ filter with 40 iterations to produce a margin of about 40 pixels. We used pixel losses, similarly as in Equation~\ref{eq:depthloss}, but assuming $\hat{y}_s$ and $y_s$ as RGB images. To increase the quality and sharpness of predicted inpainting, we also used a classic discriminator loss and a perceptual loss~\cite{johnson2016perceptual}, respectively designated by $\mathcal{L}_{disc}$ and $\mathcal{L}_{perc}$. The resulting loss for training the inpainting network is:
\begin{equation}
    \mathcal{L}_{\textit{inpaint}} = \mathcal{L}_{\textit{data}} + \alpha\mathcal{L}_{\textit{grad}} + \beta\mathcal{L}_{\textit{disc}} + \gamma\mathcal{L}_{\textit{perc}},
    \label{eq:inpaintloss}
\end{equation}
\noindent where we set $\alpha=0.1$, $\beta=0.01$, and $\gamma=0.01$. The inpainting network was also trained on our dataset with Adam and learning rate set to $0.0002$ and $0.0001$, respectively for the generator and discriminator models, for about 1 million iterations using 6 images per batch.

\subsection{Depth estimation}

We performed several experiments to evaluate the effectiveness of our depth estimation network, as well as to evaluate the proposed distillation process from MiDaS. In order to compare our \textit{lightweight} network with the \textit{teacher} network, i.e., MiDaS~\cite{Ranftl2020}, we evaluated our approach on two public depth datasets, which are described next.

The NYU-Depth V2 dataset~\cite{silberman2012indoor} is composed of 1449 pairs of images and depth maps densely annotated, from which 654 are used for evaluation. Despite the small number of samples with dense depth maps and the limited indoor images, this dataset is useful to perform ablation studies with the proposed CNN architecture, since the training process takes only one day on a single GPU.

The DIODE~\cite{diode_dataset} is a depth dataset composed of diverse high-resolution images of both indoors and outdoors scenes, paired with accurate, dense and far-ranged depth measurements. This dataset contrast itself with other depth datasets due to its multiple scene types and employment of different sensors, aiding the generalization across different domains. This dataset is useful to evaluate our proposal on diverse and precisely annotated data, specially to compare the proposed network with the \textit{teacher} model, considering the proposed knowledge distillation process.

\subsubsection{Evaluation metrics}

We evaluated our method considering classic depth estimation metrics, similarly to~\cite{eigen2014depth}. Namely, we used the \textit{threshold} metric, considering $\delta_{1.25}$, $\delta_{1.25^{2}}$, and $\delta_{1.25^{3}}$, as well as RMSE (linear) and Absolute Relative difference (REL). However, since we learn our depth estimator from MiDaS, which has scale- and shift-invariant predictions, these metrics cannot be applied directly between the ground-truth and the prediction of our model.
To allow a fair comparison, we followed the approach from~\cite{Ranftl2020} and aligned the depth predictions to the ground-truth based on the median and standard deviation. 
This \cameraready{evaluation} is very useful in our approach, since we only require normalized depth to compute our adaptive-MPI representation.
Furthermore, our dataset has no absolute ground truth depth nor camera parameters information, which prevents from evaluating depth predictions in absolute coordinates.

For new view synthesis, we use the SSIM~\cite{wang2004image} and PSNR metrics to compare predicted views with target images. Since our dataset has no paired calibrated image pairs, we compare the views produced by our adaptive slicing method with a densely sampled MPI projected to the target view point. \cameraready{Differently from \cite{Tucker_2020_CVPR}, which uses a set of points with absolute depth information to align the MPI to the ground-truth depth, we are not able to provide a fair comparison on RealEstate10K since our method is relying on relative depths only.}

\subsubsection{Network architecture}

In this part, we present a study with the depth estimation network that motivated our choices in the proposed architecture. Table~\ref{tab:ablationdepth} shows the results of our method when trained on NYU-Depth V2 only. As a baseline, we consider the same architecture as illustrated in Figure~\ref{fig:depthnetwork}, but using supervision only in the last prediction, and convolutions with striding 2 and bilinear upsampling, respectively for downscaling and upscaling feature maps.

\begin{table}[!htb]
\setlength{\tabcolsep}{0.5mm}
\centering
\caption{Results for depth estimation on NYU-Depth V2 considering variations of our depth estimation network and training strategy. +MS: multi-scale predictions, +DWT: using direct and inverse DWT, +D: pre-training on our dataset. The model \textit{lite} corresponds to the architecture from Fig.~\ref{fig:depthnetwork}, and the \textit{heavy} variation uses more filters (details in the appendix). Predictions where shifted and scaled with ground-truth based on median and standard deviation.}
\begin{tabular}{lcccc}
\toprule
Method & $\delta_{1.25}$ & RMSE & REL & Model Size \\ \midrule
Baseline model (\textit{lite}) & 0.779 & 0.531 & 0.156 & 10.8 MB \\ 
+MS & 0.793 & \underline{0.514} & \underline{0.149} & 10.8 MB \\ 
+MS +DWT & \underline{0.794} & 0.520 & 0.150 & 10.6 MB \\ 
+MS +DWT +D & 0.835 & 0.452 & 0.129 & \textbf{10.6 MB} \\ 
+MS +DWT +D (\textit{heavy}) & \textbf{0.869} & \textbf{0.416} & \textbf{0.113} & \textbf{25.2 MB} \\ \bottomrule 
\end{tabular}
\label{tab:ablationdepth}
\end{table}

\begin{figure}[!htbp]
\centering
\stackunder[5pt]{\includegraphics[width=0.16\textwidth]{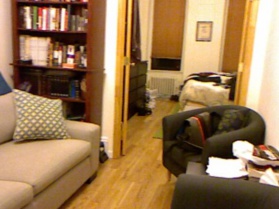}}{Input image}
\stackunder[5pt]{\includegraphics[width=0.16\textwidth]{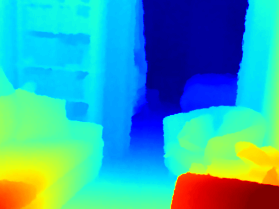}}{Ground truth}\\
\stackunder[5pt]{\includegraphics[width=0.16\textwidth]{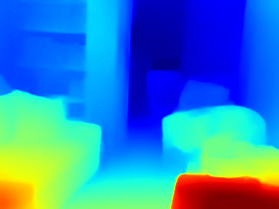}}{Baseline +MS}
\stackunder[5pt]{\includegraphics[width=0.16\textwidth]{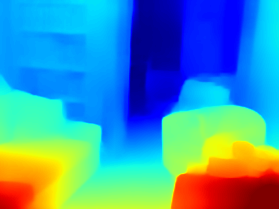}}{+MS +DWT}
\caption{Predictions on NYU-Depth V2 considering our baseline model with multiple supervisions (+MS) and the proposed architecture with DWT (\textit{lite} model version).}
\label{fig:nyuresults}
\end{figure}

We improved the baseline by 1.4\% in the $\delta_{1.25}$ score (+MS in Table~\ref{tab:ablationdepth}) just by including supervisions at multiple resolutions. This enforces the network to learn depth at different scales of the feature maps. Note that the intermediate supervisions are used only during training and, at inference time, only the last prediction is used.
Replacing strided convolutions by DWT and bilinear upsampling by inverse DWT (row +MS +DWT) improved by an additional 0.1\% and reduced the model size from 10.8 to 10.6 MB, since DWT has no trainable parameters. Despite this improvement could be considered marginal, we noticed that DWT results in sharper depth maps, as can be observed in Figure~\ref{fig:nyuresults}. This is an important factor for novel view synthesis, since most of the artifacts are produced in the borders of objects. We also show that a pre-training of the network in our dataset produces an improvement of 4.1\%. Finally, we also show the results for a \textit{heavy} version of the network, which has 25.2 MB and increases $\delta_{1.25}$ by 3.5\%.

\subsubsection{Comparison with MiDaS}

\begin{figure}[!htbp]
\centering
\includegraphics[width=0.14\textwidth]{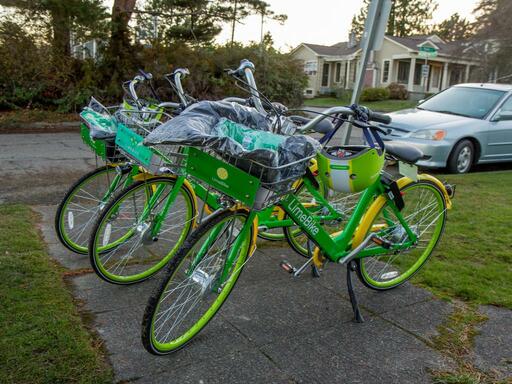}
\includegraphics[width=0.14\textwidth]{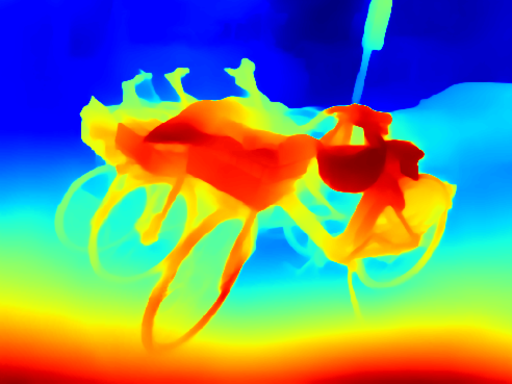}
\includegraphics[width=0.14\textwidth]{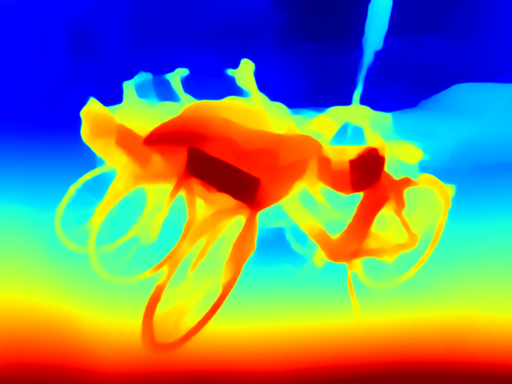}\\
\includegraphics[width=0.14\textwidth]{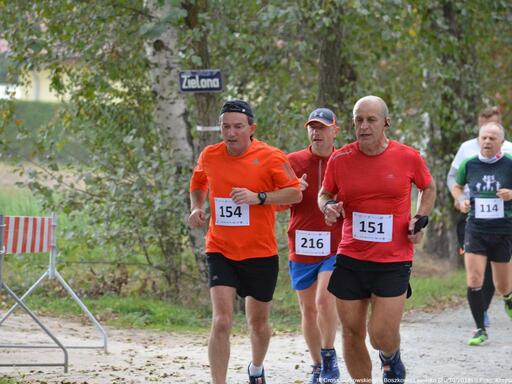}
\includegraphics[width=0.14\textwidth]{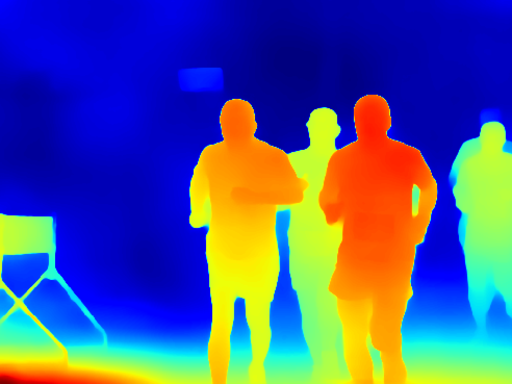}
\includegraphics[width=0.14\textwidth]{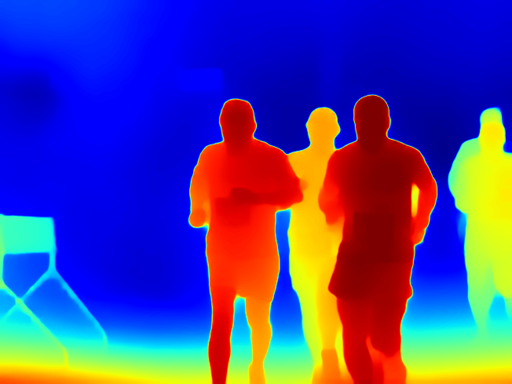}\\
\stackunder[5pt]{\includegraphics[width=0.14\textwidth]{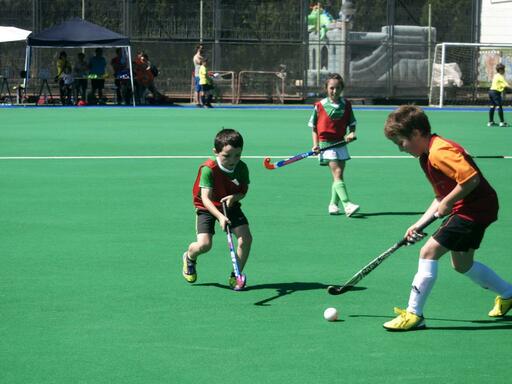}}{Input image}
\stackunder[5pt]{\includegraphics[width=0.14\textwidth]{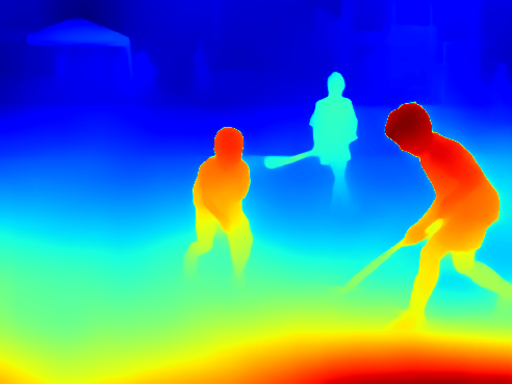}}{MiDaS}
\stackunder[5pt]{\includegraphics[width=0.14\textwidth]{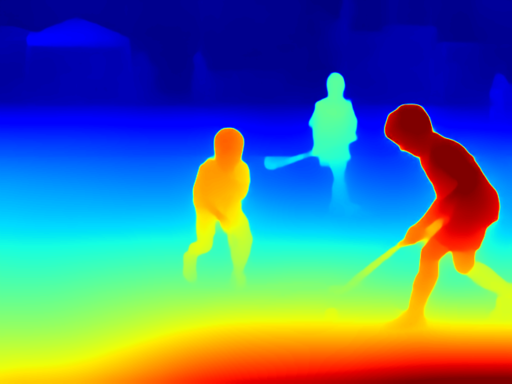}}{Ours (\textit{lite})}
\caption{Comparison of depth predictions by our \textit{lite} model and MiDaS on test samples from our dataset.}
\label{fig:comparemidas}
\end{figure}

In Table~\ref{tab:resultsmidas}, we compare our depth estimation models with MiDaS. Predictions from MiDaS and our models were scaled and shifted to the ground truth, as suggested in~\cite{Ranftl2020}. Surprisingly, our \textit{heavy} model, which has only 6.2\% of the size of MiDaS (403 MB) and is $5\times$ faster, was able to surpass the \textit{teacher} network on the DIODE dataset, while our \textit{lite} model (38 times smaller than MiDaS and $14\times$ faster) reached comparable results. We believe that this fact results from our depth distillation procedure from MiDaS, which is based on an ensemble of $10\times$ predictions (as details in Section~\ref{sec:disitllingdepth}) \cameraready{and improves the robustness of our model}. Moreover, MiDaS was trained with crops of size $384\times384$, while our model was trained on $1024\times768$, which is the native resolution of DIODE. We believe that this fact also explains our lower scores on NYU, which has lower resolution images, on which our method was not trained in. We also show in Figure~\ref{fig:comparemidas} some qualitative results of our method compared to~\cite{Ranftl2020}.

\begin{table}[!htb]
\setlength{\tabcolsep}{1.0mm}
\small
\centering
\caption{Comparison of the results from our depth estimation models on NYU-Depth V2 and DIODE datasets. On DIODE, we consider MiDaS predictions in ${384\times384}$ (native resolution) followed by a rescaling, and direct prediction in ${1024\times768}$. On NYU, all predictions are performed in ${512\times384}$. Latency was computed on an Intel Core i5 CPU at 2.60 GHz.}
\begin{tabular}{lcccccc}
\toprule
\multirow{2}{*}{Method} & \multicolumn{3}{c}{\textit{Higher is better}} & \multicolumn{3}{c}{\textit{Lower is better}} \\
\cmidrule(l){2-4} \cmidrule(l){5-7}
& $\delta_{1.25}$ & $\delta_{1.25^{2}}$ & $\delta_{1.25^{3}}$ & RMSE & REL & Latency \\ \midrule
\multicolumn{7}{c}{DIODE} \\ \midrule
MiDaS~${}_{384\times384}$ & 0.649 & 0.825 & 0.899 & 7.976 & 0.465 & 2.620 s\\
MiDaS~${}_{1024\times768}$ & 0.600 & 0.802 & 0.888 & 9.575 & 0.467 & 19.112 s \\ \midrule
\textbf{Ours} (\textit{lite}) & 0.614 & 0.812 & 0.894 & 6.420 & 0.440 & \textbf{1.308 s} \\ 
\textbf{Ours} (\textit{heavy}) & \textbf{0.684} & \textbf{0.839} & \textbf{0.902} & \textbf{5.286} & \textbf{0.385} & 3.674 s \\ \midrule 
\multicolumn{7}{c}{NYU-Depth V2} \\ \midrule
MiDaS~${}_{512\times384}$ & \textbf{0.824} & \textbf{0.951} & \textbf{0.984} & \textbf{0.550}  & \textbf{0.146} & 4.675 s \\ \midrule
\textbf{Ours} (\textit{lite}) & 0.713 &  0.925 & 0.979 & 0.640 & 0.185 & \textbf{0.339} s \\ 
\textbf{Ours} (\textit{heavy}) & 0.734 &  0.914 & 0.962 & 0.622 & 0.180 & 0.939 s \\ \bottomrule 
\end{tabular}
\label{tab:resultsmidas}
\end{table}

\begin{figure*}[!htbp]
\centering
\includegraphics[height=2.42cm]{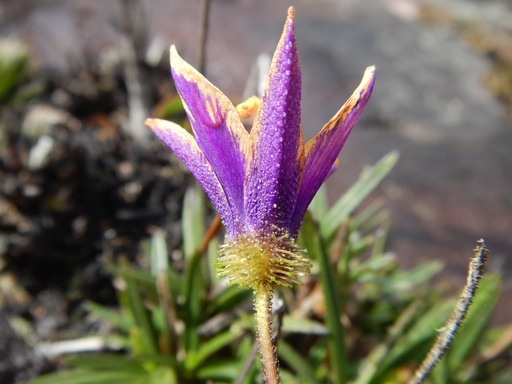}
\includegraphics[height=2.42cm]{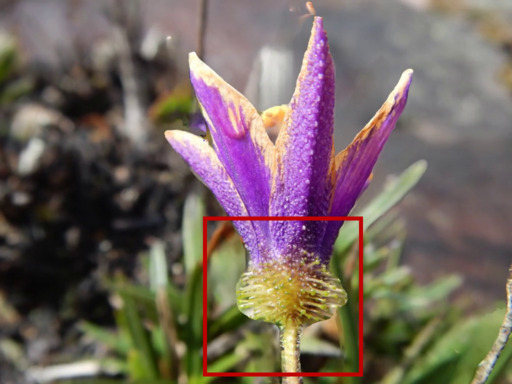}
\includegraphics[height=2.42cm]{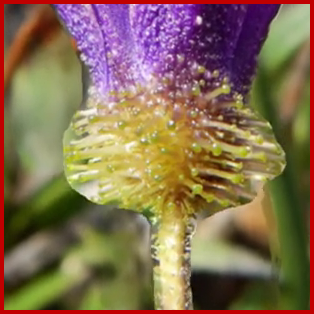}
\includegraphics[height=2.42cm]{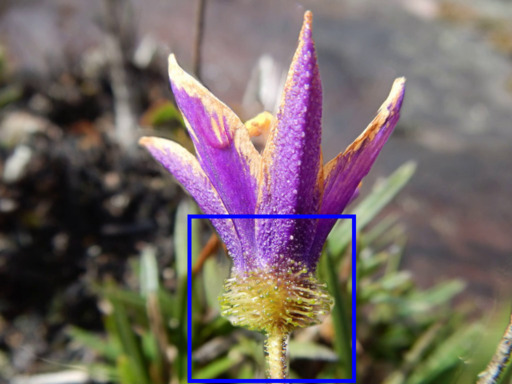}
\includegraphics[height=2.42cm]{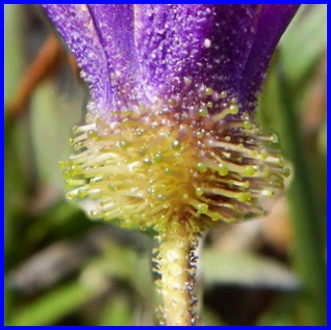}\\
\includegraphics[height=2.24cm]{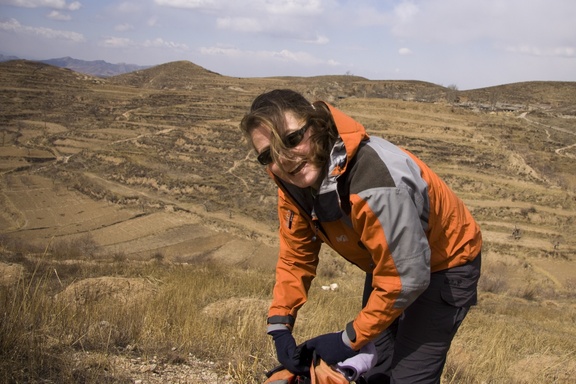}
\includegraphics[height=2.24cm]{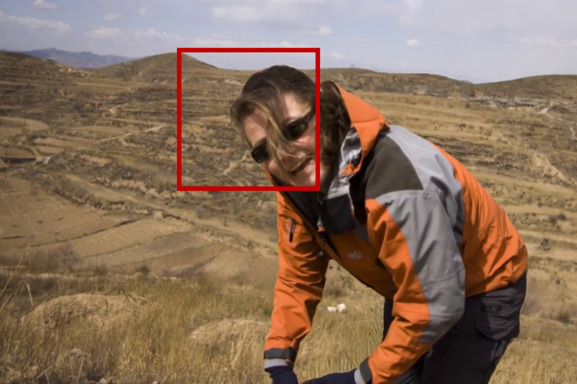}
\includegraphics[height=2.24cm]{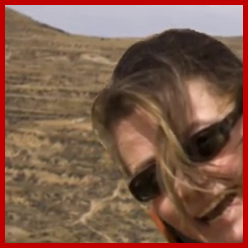}
\includegraphics[height=2.24cm]{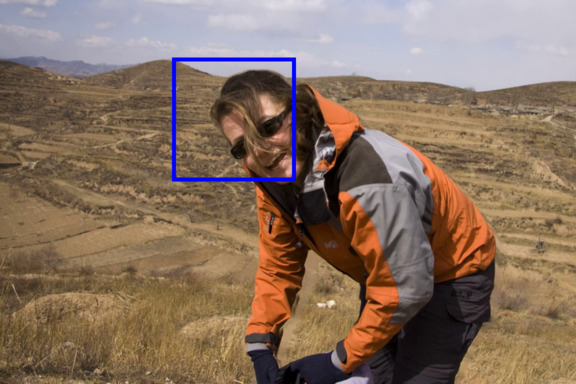}
\includegraphics[height=2.24cm]{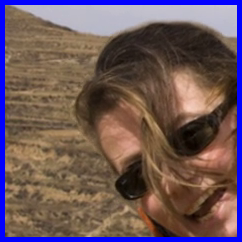}\\
\stackunder[5pt]{\includegraphics[height=2.23cm]{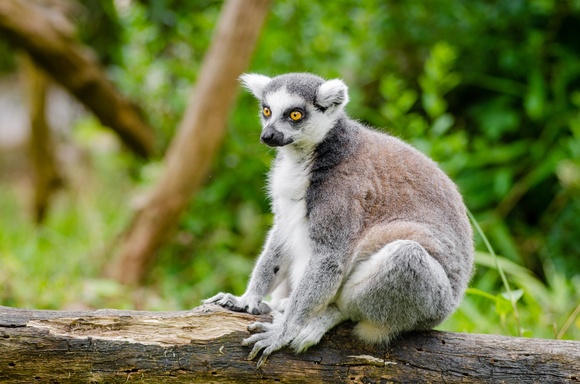}}{Input image}
\stackunder[5pt]{\includegraphics[height=2.23cm]{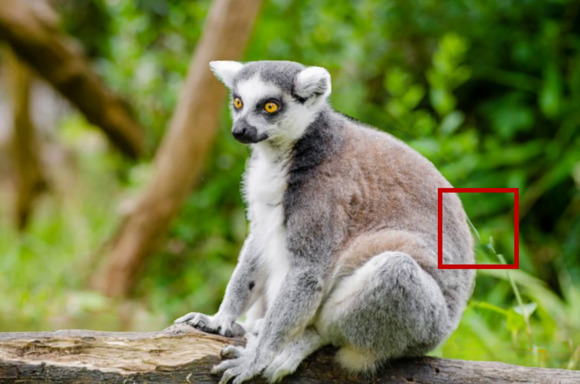}}{3D-Photography}
\stackunder[5pt]{\includegraphics[height=2.23cm]{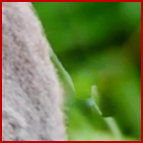}}{}
\stackunder[5pt]{\includegraphics[height=2.23cm]{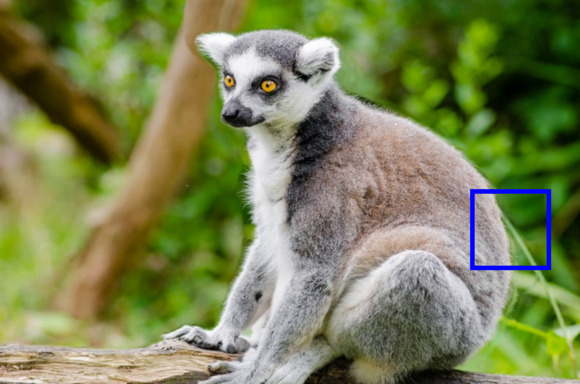}}{Ours}
\stackunder[5pt]{\includegraphics[height=2.23cm]{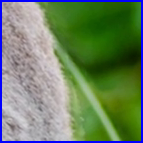}}{}
\caption{Qualitative results of our method compared to 3D-Photography on images from our dataset.}
\label{fig:compare3dphoto}
\vspace{-0.4cm}
\end{figure*}

\subsection{View synthesis}

As previously mentioned, our adaptive multiplane image slicing aims to use a small set of image planes to represent a 3D scene, while reducing the number of induced artifacts. To demonstrate the efficiency of the adaptive slicing algorithm, we used 1500 video sequences from RealEstate10K extracting a source and target image from each sequence. Varying the number of maximum planes, we used both uniform slicing and our adaptive algorithm to render the target view from the source frame using the camera transformation provided by the dataset. The results, presented in Table~\ref{tab:slicing}, show that for each number of planes the adaptive strategy outperforms the uniform slicing on both SSIM and PSNR metrics. It is important to notice that the difference between the two strategies is more significant as the number of planes gets smaller, reflecting the desired property of the adaptive slicing. Visual results of our method are shown in Figure~\ref{fig:compare3dphoto}. While 3D-Photography takes more than 400 seconds to produce an LDI running on CPU, our approach takes less than 10 seconds for $1024\times768$ images.

\begin{table}[!htb]
\setlength{\tabcolsep}{1.5mm}
\caption{Comparison between uniform and adaptive slicing using RealEstate10K. The number of maximum planes in the adaptive algorithm was varied as specified and the uniform slicing was performed with the number of planes resulting from our method.}
\centering
\begin{tabular}{clll}
\toprule
Number of planes & Method & SSIM & PSNR \\ 
\midrule
\multirow{2}{*}{4} & Uniform & 0.6710 & 18.6921 \\
& \textbf{Adaptive (ours)} & \textbf{0.6818} & \textbf{18.9200} \\
\midrule
\multirow{2}{*}{8} & Uniform & 0.6778 & 18.8438 \\
& \textbf{Adaptive (ours)} & \textbf{0.6836} & \textbf{18.9725} \\
\midrule
\multirow{2}{*}{16} & Uniform & 0.6804 & 18.9034 \\
& \textbf{Adaptive (ours)} & \textbf{0.6841} & \textbf{18.9874} \\
\bottomrule
\end{tabular}
\label{tab:slicing}
\end{table}

\subsubsection{Subjective user study}

In addition to the objective evaluation carried out, we also conducted a subjective study with users through 25 images chosen randomly from the test samples of our dataset to evaluate and compare our method with 3D-Photography~\cite{shih20203d} according to human perception. We generated videos with equivalent motion for both methods, placing them side by side in a single video. Three types of movement were used based on those provided by the 3D-Photography code. Each video was generated with 180 frames of size $1024 \times 768$ and 30 frames per second.

We asked users to evaluate two aspects of the videos: (a) motion quality, in order to check whether the motion is natural and (b) amount of noise or artifacts present in the video. For both cases, we use the Absolute Category Rating, with 1 indicating poor and 5 indicating excellent. In addition, we allow the user to optionally comment on each assessment. We obtained, for each of the five video samples, the respective number of responses from users: 22, 18, 17, 15, and 15. For each of the videos, we calculated their average for the values of the motion and frame quality. Then, we applied the min-max normalization to change the range of averages from [1,5] to [0,100]. We also count, for each video, the number of times that someone assigned an equal or better rating to our method. Table~\ref{tab:user_study} shows the Mean Opinion Score (MOS) and the percentage of times that our videos scored equal or higher for both aspects evaluated.

\begin{table}[!htb]
\setlength{\tabcolsep}{2.5mm}
\caption{Comparison between our method and 3D-Photography according to users' perception.}
\centering
\begin{tabular}{lll}
\toprule
MOS & Motion Quality & Frame Quality \\ 
\midrule
3D-Photography & 70.24 $\pm$ 7.36 & 72.71 $\pm$ 8.61 \\
Ours           & 65.07 $\pm$ 8.82 & 64.49 $\pm$ 14.23 \\
\midrule
Ours (\% of $\geq$) & 63.85 $\pm$ 12.94 & 60.37 $\pm$ 23.17 \\
\bottomrule
\end{tabular}
\label{tab:user_study}
\end{table}

From these results, we can notice that 3D-Photography had some superiority in MOS. However, the difference is not substantial, being comparable by the standard deviation. In addition, our method scored higher than or equal to more than 60\% of the responses in both criteria, which is even more relevant considering that our method requires significantly less memory and computational processing.

\section{Conclusions}\label{sec:conclusions}

In this paper, we described a novel method to estimate an adaptive multiplane image representation from a single image that allows novel view synthesis with low computational requirements. The main contribution presented in our work is a new algorithm that produces a variable set of image planes depending on the depth of the scene. In addition, we proposed a distillation technique that allows our lightweight CNN architecture to achieve depth estimation results comparable to a state-of-the-art model while requiring only 2.7\% of the original model's number of parameters (a reduction of 38 times in the model size). Therefore, our method is capable of producing an adaptive-MPI for high-resolution images in less than 10 seconds (about 44 times faster than 3D-Photography), resulting in an efficient method for new high-quality view generation.


{
  \small
  \bibliographystyle{ieee}
  \bibliography{references}
}

\end{document}